\documentclass[10pt,twocolumn,letterpaper]{article}

\usepackage[pagenumbers]{cvpr} 

\definecolor{citecolor}{HTML}{0071BC}
\definecolor{linkcolor}{HTML}{ED1C24}

\usepackage[
    pagebackref,
    breaklinks,
    colorlinks,
    citecolor=citecolor,
    linkcolor=linkcolor,
    urlcolor=gray
]{hyperref}

\usepackage{bm}
\usepackage{colortbl}
\usepackage{multirow}
\usepackage{xcolor}
\usepackage{graphicx}
\usepackage{cuted}
\usepackage{caption} 

\usepackage{colortbl}
\newcommand{\tablestyle}[2]{%
    \setlength{\tabcolsep}{#1}%
    \renewcommand{\arraystretch}{#2}%
    \centering\footnotesize
}

\renewcommand{\paragraph}[1]{\vspace{1.25mm}\noindent\textbf{#1}}

\definecolor{gainGreen}{RGB}{0,140,0}
\definecolor{lossRed}{RGB}{215,0,0}

\newcommand{\gain}[1]{%
    \raisebox{.12ex}{\textcolor{gainGreen}{\tiny\bfseries (#1)}}%
}
\newcommand{\loss}[1]{%
    \raisebox{.12ex}{\textcolor{lossRed}{\tiny\bfseries (#1)}}%
}

\newlength\savewidth
\newcommand\shline{\noalign{\global\savewidth\arrayrulewidth
  \global\arrayrulewidth 1pt}\hline\noalign{\global\arrayrulewidth\savewidth}}

\newcommand\midline{\noalign{\global\savewidth\arrayrulewidth
  \global\arrayrulewidth 0.5pt}\hline\noalign{\global\arrayrulewidth\savewidth}}

\def\confName{CVPR}
\def\confYear{2027}

\title{Prune to Expand: Training Prompt Enhancers with Less to Imitate}

\author{
	Tianhe Wu$^{1}$, Qilin Sun$^{2}$, and 
	Kede Ma$^{1}$\thanks{Corresponding author.}\\
	$^1$City University of Hong Kong \ 
	$^2$The Chinese University of Hong Kong (Shenzhen)\\
	{\tt\small tianhewu-c@my.cityu.edu.hk}\; {\tt\small sunqilin@cuhk.edu.cn}\; {\tt\small kede.ma@cityu.edu.hk}\\
}

\begin{document}
\setcounter{footnote}{1}
\maketitle

\begin{strip}
    \centering
    \vspace{-7em}
    \makebox[\linewidth][c]{%
        \includegraphics[width=0.95\linewidth]{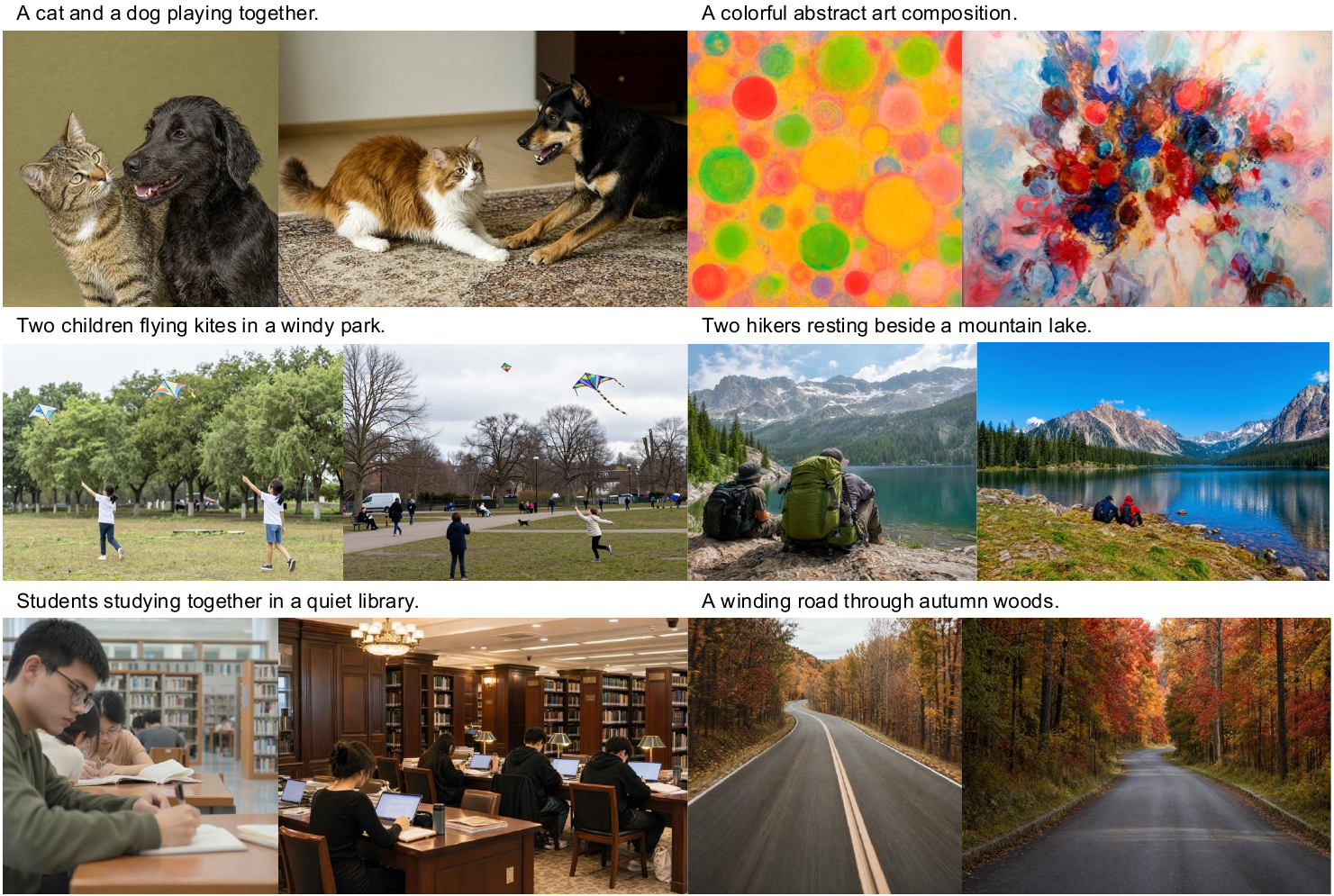}%
        \vphantom{\includegraphics[width=\linewidth]{figures/teaser.pdf}}%
    }
\captionof{figure}{
\textbf{Qualitative results of Prune to Expand (P2E).}
For each prompt, the left image is generated from the original short prompt, while the right image is generated from the prompt enhanced by the PE model fine-tuned on P2E-refined supervision. By enriching short prompts with intent-aligned visual details, P2E enables more vibrant colors, vivid scenes, and aesthetically balanced compositions.}
    \label{fig:teaser}
\end{strip}

\begin{abstract}
Prompt enhancement (PE) refines concise user prompts into detailed descriptions to improve the compositional aesthetics of generated images while preserving user intent. A common practice constructs supervised fine-tuning (SFT) pairs by deriving short, user-like prompts from dense captions, implicitly assuming that captions that improve image generation also provide effective supervision. In this paper, however, we uncover a counterintuitive finding: PE models trained on caption pairs with substantial generation gains perform comparably to those trained on pairs with only marginal gains. We argue that dense captions can impose redundant supervision by requiring PE models to reproduce image-specific details weakly related to user intent. Motivated by this discrepancy, we introduce \textbf{Prune to Expand} (P2E), a simple text-only method that refines dense-caption supervision by selectively removing redundant sentences under a per-example deletion bound while preserving explicit user requirements and informative visual structure. Extensive experiments under identical training configurations demonstrate that P2E improves alignment with human preferences and visual quality over conventional dense-caption SFT across diverse image categories and visual generators.
\end{abstract}

\section{Introduction}
\label{sec:intro}
Visual generative models have become increasingly capable of translating textual descriptions into diverse, high-quality visual content, supporting a growing range of everyday creative workflows~\cite{seedream2025seedream, seedance2026seedance, team2025hunyuanimage, zhao2026qwen, esser2024scaling}. These advances have broadened access to visual creation and made natural language an increasingly important interface for expressing and realizing creative intent.

Extensive research has advanced visual quality~\cite{kingma2013auto, goodfellow2014generative, song2020score, ho2020denoising, lipman2022flow}, instruction following~\cite{xu2023imagereward, kirstain2023pick, li2026qwen}, generation efficiency~\cite{song2023consistency, song2023improved, yin2024one, geng2026mean, wu2026qwen}, and alignment with human perception~\cite{liu2026flow, he2025tempflow, zheng2026diffusionnft, liu2026beyond}. Despite this progress, users typically express their intentions through \textit{concise}, high-level prompts that leave visual details and compositional choices underspecified. Consequently, generated images may exhibit satisfactory visual quality while lacking the compositional coherence and aesthetic refinement needed to fully realize the potential of the user's request~\cite{zhang2026qwen}.

To bridge this gap, prompt enhancement (PE) enriches concise user instructions with meaningful visual details while preserving their underlying intent, guiding visual generators toward more vivid, compositionally harmonious, and aesthetically compelling images~\cite{hao2023optimizing, datta2024prompt, wang2026promptenhancer, zhang2026qwen, lin2026pe}. A common training strategy constructs paired short prompts and dense captions for supervised fine-tuning (SFT) of pretrained large language models (LLMs)~\cite{manas2024improving, yeh2026tipo, cao2023beautifulprompt, yao2024promptcot, zhang2026qwen}. This practice rests on an intuitive assumption: dense captions that improve image generation also provide effective supervision for learning prompt enhancement.

In this paper, however, we challenge this assumption with a counterintuitive empirical finding: \textit{PE models trained on caption pairs yielding substantial generation gains perform comparably to those trained on pairs yielding only marginal gains}, as shown in \autoref{tab:motivation}. We argue that dense captions can impose \textit{redundant supervision} by requiring PE models to reproduce image-specific details weakly related to user intent. Such details may benefit direct image generation without providing equally useful learning signals for prompt enrichment. Further pruning experiments suggest that some of this content can be removed while largely preserving downstream performance, whereas unconstrained deletion can discard valuable supervision.

Motivated by these observations, we revisit supervision design for PE and introduce \textbf{Prune to Expand} (P2E), a simple text-only method that selectively removes redundant sentences while preserving explicit user requirements and informative visual structure. Given a short prompt and its paired dense caption, P2E employs an LLM to construct progressively pruned candidates, keeping the wording and order of retained sentences unchanged. It then selects the candidate with the largest token removal ratio within a predefined per-example deletion bound, explicitly controlling the extent of supervision pruning. This refinement operates entirely offline, requiring \textit{no image generation or visual scoring and introducing no additional inference overhead}.

Experiments across diverse image categories and pretrained visual generators demonstrate that P2E improves alignment with human preferences and visual quality over conventional dense-caption SFT under identical training configurations. More broadly, our findings advocate for a simple principle: \textit{prune to expand}. In an era of increasingly capable foundation models, we hope this work highlights the importance of rethinking supervision design, shifting the focus from maximizing the richness of training targets to identifying what truly matters for effective learning.

\section{Related Work}
\label{sec:related_work}
This section reviews two closely related research directions: prompt enhancement for visual generation and caption curation and compression, highlighting their respective contributions and connections to our supervision design.

\paragraph{Prompt Enhancement.}
Prompt enhancement bridges the gap between concise user requests and effective textual conditioning for visual generation. Promptist~\cite{hao2023optimizing} combines SFT with reinforcement learning to improve aesthetics while preserving user intent. PAE~\cite{mo2024dynamic} extends prompt optimization to word weights and their injection timesteps during diffusion. Another line of work learns suitable prompt distributions: PromptCoT~\cite{yao2024promptcot} aligns prompts with descriptions associated with high-quality images, Datta \etal~\cite{datta2024prompt} learn expansions that encourage both aesthetics and diversity, and TIPO~\cite{yeh2026tipo} samples detailed prompts from a target text distribution. Chen \etal~\cite{chen2026scaling} study how structured text conditioning scales and train a prompter to add semantic and geometric details to user requests. Visual feedback provides a complementary source of guidance, through iterative prompt revision in Idea2Img~\cite{yang2024idea2img} and fine-grained alignment rewards in PromptEnhancer~\cite{wang2026promptenhancer}. AP-Adapter~\cite{fu2024apadapter} further studies transfer to unseen generators by adapting prompt representations across models. Most closely related to our setting, Qwen-Image-PE~\cite{zhang2026qwen} constructs supervision from dense captions of well-designed images through reverse distillation, followed by reinforcement learning with design rewards. While prior work advances prompt construction, feedback, and generator adaptation, we show that captions yielding larger generation gains need not provide better PE supervision.

\paragraph{Caption Pruning.}
Caption curation has been widely studied in vision-language learning.
BLIP~\cite{li2022blip} bootstraps captions through generation and filtering, CapsFusion~\cite{yu2024capsfusion} combines information from raw and synthetic captions, and ShareGPT4V~\cite{chen2024sharegpt4v} demonstrates the benefits of detailed captions for multimodal alignment. These findings highlight the value of better captions but leave open which details should supervise prompt enrichment. Separately, sentence compression and summarization have long addressed textual redundancy. Deletion-based compression~\cite{filippova2015sentence} learns which words to retain, while Swell-and-Shrink~\cite{wang2019swell} condenses regional descriptions into an image caption. For language models, Selective Context~\cite{li2023compressing} and LLMLingua~\cite{jiang2023llmlingua} reduce input redundancy using language-model information estimates. LongLLMLingua~\cite{jiang2024longllmlingua} makes compression query-aware, and RECOMP~\cite{xu2024recomp} selects or summarizes retrieved content according to its utility for downstream prediction. Whereas prior work aims to improve caption quality or condense text, P2E uses each short prompt to select sentences from its paired dense caption as supervision for prompt enhancement.

\begin{table}[t]
    \centering
    \tablestyle{2.5pt}{1.2}
    \footnotesize
    \caption{\textbf{Generation gains} versus downstream PE performance. The upper section compares generation from short and dense prompts in $\mathcal{D}_M$ and $\mathcal{D}_H$, while the lower section evaluates the PE models on $200$ held-out prompts. ``w/o PE'' uses short prompts, while ``Raw PE'' uses the pretrained PE model without fine-tuning. Colored annotations show differences from the preceding row. All images are generated by Qwen-Image-2.1-Turbo.}
    \vspace{-.3em}
    \begin{tabular}{l|l@{\,}ll@{\,}l|l@{\,}ll@{\,}l}
         \multirow{2}{*}{\textbf{Method}}
         & \multicolumn{4}{c|}{\textbf{Preference}}
         & \multicolumn{4}{c}{\textbf{Visual Quality}} \\
         & \multicolumn{2}{l}{HPS v2$\uparrow$}
         & \multicolumn{2}{l|}{PicS$\uparrow$}
         & \multicolumn{2}{l}{MIQA$\uparrow$}
         & \multicolumn{2}{l}{VQ-R1$\uparrow$} \\
        \shline

        \rowcolor[gray]{0.9}
        \multicolumn{9}{l}{\textit{Training set (direct generation)}} \\

        \multicolumn{9}{l}{\hspace{.2em}\textbf{Marginal-gain subset} ($\mathcal{D}_M$)} \\
        \hspace{.4em}Short prompt
            & 29.58 &
            & 22.50 &
            & 0.630 &
            & 4.005 & \\
        \hspace{.4em}Dense caption
            & 29.71 & \gain{+0.13}
            & 22.45 & \loss{-0.05}
            & 0.658 & \gain{+0.028}
            & 4.010 & \gain{+0.005} \\
        \hline

        \multicolumn{9}{l}{\hspace{.2em}\textbf{High-gain subset} ($\mathcal{D}_H$)} \\
        \hspace{.4em}Short prompt
            & 26.15 &
            & 21.93 &
            & 0.640 &
            & 3.990 & \\
        \hspace{.4em}Dense caption
            & 29.35 & \gain{+3.20}
            & 22.45 & \gain{+0.52}
            & 0.672 & \gain{+0.032}
            & 3.990 & \gain{+0.000} \\
        \midline

        \rowcolor[gray]{0.9}
        \multicolumn{9}{l}{\textit{Held-out test set}} \\
        \hspace{.4em}Baseline (w/o PE)
            & 26.21 &
            & 22.69 &
            & 0.559 &
            & 4.000 & \\
        \hspace{.4em}Raw PE
            & 26.73 & \gain{+0.52}
            & 22.63 & \loss{-0.06}
            & 0.556 & \loss{-0.003}
            & 4.015 & \gain{+0.015} \\
        \hline
        \hspace{.4em}Marginal-gain PE
            & 27.13 &
            & 22.76 &
            & 0.573 &
            & 4.050 & \\
        \hspace{.4em}High-gain PE
            & 27.27 & \gain{+0.14}
            & 22.81 & \gain{+0.05}
            & 0.575 & \gain{+0.002}
            & 4.050 & \gain{+0.000} \\
    \end{tabular}
    \vspace{-.8em}
    \label{tab:motivation}
\end{table}

\section{Do Generation Gains Improve PE Training?}
\label{sec:motivation}
PE models are commonly trained to imitate dense captions that improve image generation. However, effective generation captions need not provide effective supervision. We show that larger generation gains offer \textit{limited} training benefits, while removing redundant supervision requires careful control over content and deletion extent.

\subsection{Measuring Generation Gains}
\label{sec:formulation}
We first examine this distinction by constructing two SFT datasets with contrasting generation gains. Let $\bm{x}_i$ and $\bm{y}_i$ denote a short prompt and its corresponding dense caption, respectively. Given a fixed visual generator $g_{\bm{\theta}}$ and an evaluator $q$, we define the generation gain as
\begin{equation}
    \Delta q_i :=
    q(g_{\bm{\theta}}(\bm{y}_i), \bm{x}_i)
    - q(g_{\bm{\theta}}(\bm{x}_i), \bm{x}_i),
    \label{eq:generation_gain}
\end{equation}
where $q$ evaluates image quality conditioned on the original prompt $\bm{x}_i$. A larger $\Delta q_i$ indicates greater generation gains from dense captions. We define the average gain of a dataset $\mathcal{D}$ as $\overline{\Delta q}_{\mathcal{D}} = \frac{1}{|\mathcal{D}|}\sum_{i \in \mathcal{D}}\Delta q_i$. Accordingly, we construct two equally sized SFT datasets: a high-gain subset $\mathcal{D}_H$ and a marginal-gain subset $\mathcal{D}_M$, each containing 10K original short prompts paired with their corresponding detailed dense captions for SFT.

\subsection{Testing High-gain Supervision}
\label{sec:motivation_exp}

\paragraph{Experimental Setups.}
\label{sec:exp_setup_demo}
We first curate a diverse image collection $\mathcal{D}$ from ImageNet 1024$\times$1024~\cite{deng2009imagenet}. For each image, we employ Qwen3-VL-32B-Instruct~\cite{bai2025qwen3} to generate a dense caption, subsequently degraded into a shorter, user-like prompt using the same model conditioned on the image and caption\footnote{The short prompts are synthesized to mimic natural user inputs, rather than derived by directly shortening dense captions.}. We adopt Qwen-Image-2.1-Turbo~\cite{zhao2026qwen} at its native 2K resolution as the fixed visual generator $g_{\bm{\theta}}$ to synthesize images from both prompts. Using HPS v2~\cite{wu2023human}, we quantify their generation gain $\Delta q_i$ relative to the original short prompt, selecting the 10K pairs with the highest and lowest gains to form $\mathcal{D}_H$ and $\mathcal{D}_M$, respectively.

We then independently fine-tune two PE models based on Qwen3.5-9B~\cite{yang2025qwen3} with identical initialization and training configurations, differing only in their supervision datasets. Specifically, given an SFT dataset $\mathcal{D}$, we minimize the autoregressive objective:
\begin{equation}
    \ell_{\mathrm{SFT}}(\bm{\phi})
    \triangleq -\mathbb{E}_{\mathcal{D}}
    \left[\sum_{t=1}^{T}
    \log p_{\bm{\phi}}(\bm{y}_t\mid\bm{x},\bm{y}_{<t})\right],
    \label{eq:sft_loss}
\end{equation}
where $p_{\bm{\phi}}$ denotes the PE model parameterized by $\bm{\phi}$, and $T$ is the length of the target caption. The loss is computed only over target tokens, with the original short prompt serving as conditioning context. Both models are evaluated on the same $200$ held-out user prompts using HPS v2, PickScore (PicS)~\cite{kirstain2023pick}, MANIQA (MIQA)~\cite{yang2022maniqa}, and VisualQuality-R1 (VQ-R1)~\cite{wu2025visualquality}. The former two assess prompt-conditioned human preferences, while the latter two measure prompt-independent perceptual image quality. Further details are provided in the supplementary material.

\paragraph{High Gains, Marginal Benefits.}
As shown in \autoref{tab:motivation}, $\mathcal{D}_H$ achieves an average generation gain of $\overline{\Delta q}_{\mathcal{D}_H}=3.20$, approximately 25 times that of $\mathcal{D}_M$ ($\overline{\Delta q}_{\mathcal{D}_M}=0.13$). Despite this substantial difference, the resulting PE models exhibit comparable downstream performance across all four evaluation metrics, with only marginal improvements from high-gain supervision. This counterintuitive finding suggests that substantially greater generation gains do not necessarily translate into more effective supervision for PE.

\begin{table}[t]
    \centering
    \tablestyle{4pt}{1.2}
    \footnotesize
    \caption{\textbf{Probing supervision redundancy} through caption pruning. We compare two preliminary pruning strategies against the baseline trained on unpruned dense captions from $\mathcal{D}_H$.}
    \vspace{-.3em}
    \begin{tabular}{l|cc|cc}
         \multirow{2}{*}{\textbf{Method}}
         & \multicolumn{2}{c|}{\textbf{Preference}}
         & \multicolumn{2}{c}{\textbf{Visual Quality}} \\
         & HPS v2$\uparrow$ & PicS$\uparrow$
         & MIQA$\uparrow$ & VQ-R1$\uparrow$ \\
        \shline
        Baseline (high-gain PE)
            & 27.27
            & 22.81
            & 0.575
            & 4.050 \\ \hline
        + LLM-only pruning
            & 26.46
            & 22.53
            & 0.570
            & 4.015 \\
        + Visual-guarded pruning
            & 27.59
            & 22.77
            & 0.574
            & 4.035 \\
    \end{tabular}
    \vspace{-.8em}
    \label{tab:probing}
\end{table}

\subsection{Probing Supervision Redundancy}
\label{sec:pruning_probe}
We attribute this discrepancy to redundant supervision in dense captions, where image-specific details may benefit visual generation but offer limited learning signals for PE models conditioned on short prompts. To probe this argument, we conduct two pruning experiments on $\mathcal{D}_H$, with and without visual feedback to validate deletions.

\paragraph{Probing via Pruning.}
In both experiments, we employ Qwen3.5-9B~\cite{yang2025qwen3} to selectively remove image-specific details weakly related to the original short prompt, producing a pruned caption $\tilde{\bm{y}}_i$ from $\bm{y}_i$. We explore two strategies: (i) LLM-only pruning directly uses the pruned captions as SFT targets without further verification; (ii) Visual-guarded pruning introduces a visual quality constraint, accepting a pruned caption only if
\begin{equation}
    q(g_{\bm{\theta}}(\bm{y}_i), \bm{x}_i)
    - q(g_{\bm{\theta}}(\tilde{\bm{y}}_i), \bm{x}_i)
    \leq \varepsilon,
    \label{eq:pruning_constraint}
\end{equation}
where $q$ denotes HPS v2~\cite{wu2023human} and $\varepsilon=0.1$. Otherwise, the original caption is retained. We independently fine-tune PE models on both datasets under the same training configurations as in \autoref{sec:motivation_exp}.

\paragraph{Evidence of Removable Redundancy.}
As shown in \autoref{tab:probing}, LLM-only pruning substantially degrades downstream PE performance, whereas visual-guarded pruning largely preserves the original performance despite removing redundant content. These results suggest that dense captions indeed contain \textit{removable redundancy}, but unconstrained deletion may discard valuable supervision.

\begin{figure}[t]
    \centering
    \includegraphics[width=0.9\linewidth]{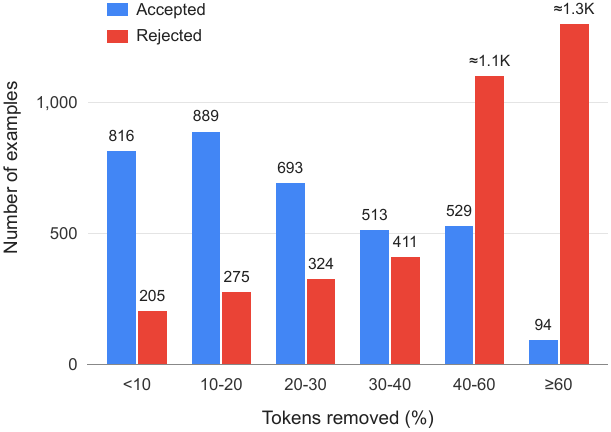}
    \vspace{-.3em}
    \caption{\textbf{Effect of pruning extent.} Distribution of accepted and rejected pruning proposals under (ii) visual-guarded pruning on $\mathcal{D}_H$, grouped by the proportion of removed tokens, showing that aggressive deletions are more frequently rejected.}
    \vspace{-.8em}
    \label{fig:length}
\end{figure}

\paragraph{Pruning Extent Matters.}
Unlike LLM-only pruning, visual-guarded pruning explicitly rejects proposals that fail visual verification. To understand what characterizes these rejected proposals, we examine their relationship with deletion extent in \autoref{fig:length}. We observe that moderate deletions are more likely to be accepted, whereas aggressive deletions are predominantly rejected. This suggests that excessive pruning may discard essential supervision, highlighting the importance of \textit{controlling} deletion extent.

\begin{figure*}[t]
    \centering
    \includegraphics[width=0.98\linewidth]{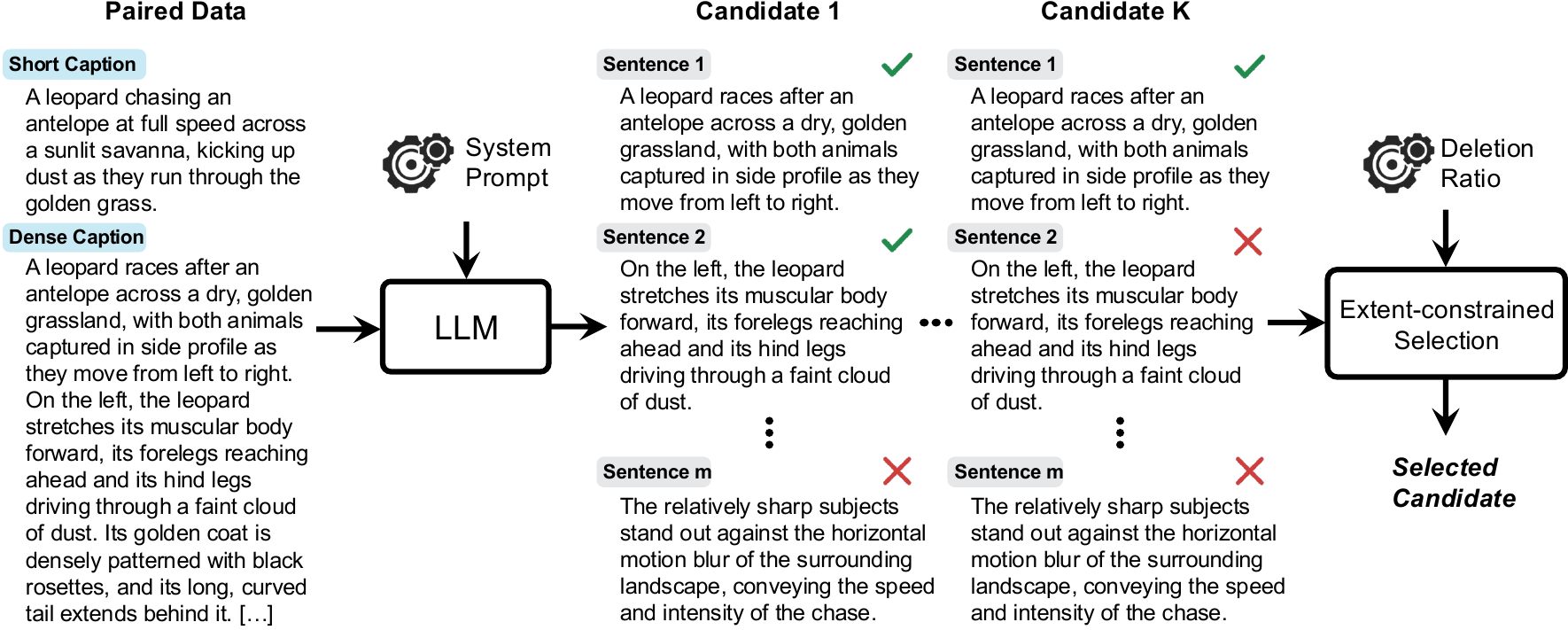}
    \vspace{-.3em}
    \caption{\textbf{Overview of Prune to Expand (P2E).} Given a short prompt and its paired dense caption, an LLM generates $K$ nested candidates by progressively removing sentences weakly related to user intent. Retained sentences remain unchanged and in their original order. Extent-constrained selection then chooses the candidate with the largest token deletion ratio within the per-example bound $\tau$ as the refined supervision target for PE. Green checks and red crosses indicate retained and removed sentences, respectively. The pruning process is entirely text-only, requiring no image generation or visual scoring.}
    \vspace{-.8em}
    \label{fig:method}
\end{figure*}

\section{Prune to Expand}
\label{sec:p2e}
Driven by the above analysis, we propose \textbf{Prune to Expand} (P2E), a simple and effective text-only method that selectively removes redundant supervision from dense captions while preserving user intent and controlling deletion extent, without costly visual verification.

\subsection{Pruning Principles}

\paragraph{Sentence-level Pruning.}
PE aims to enrich user prompts with meaningful visual details beyond their explicit content. Compared with word-level pruning, which may fragment coherent descriptions and discard valuable visual details, sentence-level pruning better preserves semantic integrity and retains meaningful, intent-aligned elaborations of visual details and compositional structure. We therefore remove sentences weakly related to user intent, while retaining those conveying explicit requirements or meaningful visual structure without altering their wording or order.

\paragraph{Pruning Extent Constraint.}
Identifying redundant content alone is insufficient, since excessive deletion may discard valuable supervision, as shown in \autoref{fig:length}. Moreover, sentence-level pruning inevitably introduces discrete changes in caption length, making precise control of deletion extent difficult. We therefore impose a per-example deletion bound $\tau$ on the proportion of removed tokens, rather than enforcing an exact deletion ratio. The deletion bound provides a lightweight control on pruning extent, while semantic relevance is assessed by the LLM during candidate construction.

\subsection{Computational Procedure of P2E}
\label{sec:p2e_computation}
As illustrated in \autoref{fig:method}, P2E operates on paired user prompts and dense captions. Given a pair $(\bm{x}_i,\bm{y}_i)$, we employ an LLM conditioned on both texts to generate $K$ nested pruning candidates $\{\tilde{\bm{y}}_i^{(k)}\}_{k=1}^{K}$ with progressively decreasing token counts, \ie, $|\tilde{\bm{y}}_i^{(1)}|_{\mathrm{tok}} > \cdots > |\tilde{\bm{y}}_i^{(K)}|_{\mathrm{tok}}$, where $|\cdot|_{\mathrm{tok}}$ denotes the token count. Each candidate removes more intent-irrelevant sentences while preserving the original wording and order of retained sentences.

Since sentence-level pruning introduces discrete changes in caption length, we compute the actual deletion ratio of each candidate $\tilde{\bm{y}}_i^{(k)}$ as
\begin{equation}
    \tau_i^{(k)} =
    1-\frac{|\tilde{\bm{y}}_i^{(k)}|_{\mathrm{tok}}}
    {|\bm{y}_i|_{\mathrm{tok}}}.
\end{equation}
Given the deletion bound $\tau$, we select the most aggressively pruned candidate satisfying the constraint:
\begin{equation}
    k_i^\star =
    \underset{k\in\{0,\ldots,K\}}{\arg\max}
    \;\tau_i^{(k)}
    \quad \text{s.t.}\ \tau_i^{(k)}\leq\tau,
    \label{eq:p2e_selection}
\end{equation}
where $\tilde{\bm{y}}_i^{(0)}=\bm{y}_i$ and $\tau_i^{(0)}=0$ represent the original dense caption. Including this unpruned candidate ensures a feasible solution even when all pruning candidates exceed the deletion bound. 

The selected captions form the refined SFT dataset $\mathcal{D}_{\mathrm{P2E}} =
\{(\bm{x}_i,\tilde{\bm{y}}_i^{(k_i^\star)})\}_{i=1}^{N}$, where $N$ is the number of training pairs. We fine-tune the PE model on $\mathcal{D}_{\mathrm{P2E}}$ using the autoregressive objective in \autoref{eq:sft_loss}. P2E operates entirely offline without image generation or visual evaluation, introducing no additional inference overhead.

\begin{table*}[t]
    \centering
    \tablestyle{2.2pt}{1.2}
    \scriptsize
    \caption{\textbf{Main results across four categories and four visual generators.} Each category contains $200$ held-out prompts. Both SFT variants use identical training settings but different supervision targets: unpruned versus P2E-refined captions. P2E generally outperforms conventional SFT across categories and generators, highlighting the effectiveness and generalizability of supervision pruning. The best and second-best results are highlighted in \textbf{bold} and \underline{underline}, respectively.}
    \label{tab:main_results}
    \vspace{-.3em}

    \begin{tabular}{l|*{3}{cc|cc|}cc|cc}
         \multirow{3}{*}{\textbf{Method}}
         & \multicolumn{4}{c|}{\textbf{Landscape}}
         & \multicolumn{4}{c|}{\textbf{Design}}
         & \multicolumn{4}{c|}{\textbf{Portrait}}
         & \multicolumn{4}{c}{\textbf{General}} \\

         & \multicolumn{2}{c|}{\textbf{Preference}}
         & \multicolumn{2}{c|}{\textbf{Visual Quality}}
         & \multicolumn{2}{c|}{\textbf{Preference}}
         & \multicolumn{2}{c|}{\textbf{Visual Quality}}
         & \multicolumn{2}{c|}{\textbf{Preference}}
         & \multicolumn{2}{c|}{\textbf{Visual Quality}}
         & \multicolumn{2}{c|}{\textbf{Preference}}
         & \multicolumn{2}{c}{\textbf{Visual Quality}} \\

         & HPS v2$\uparrow$ & PicS$\uparrow$ & MIQA$\uparrow$ & VQ-R1$\uparrow$
         & HPS v2$\uparrow$ & PicS$\uparrow$ & MIQA$\uparrow$ & VQ-R1$\uparrow$
         & HPS v2$\uparrow$ & PicS$\uparrow$ & MIQA$\uparrow$ & VQ-R1$\uparrow$
         & HPS v2$\uparrow$ & PicS$\uparrow$ & MIQA$\uparrow$ & VQ-R1$\uparrow$ \\
        \shline

        \rowcolor[gray]{0.9}
        \multicolumn{17}{l}{\textit{Qwen-Image-2.1-Turbo}} \\

        \hspace{.4em}w/o PE
            & 24.47 & 22.75 & 0.636 & \underline{4.005}
            & 25.97 & 22.29 & 0.613 & \underline{4.000}
            & 30.14 & 23.25 & 0.553 & 4.000
            & 26.81 & 22.69 & 0.572 & 4.010 \\

        \hspace{.4em}Raw PE
            & \underline{24.72} & \underline{22.89} & 0.631 & 3.990
            & 26.31 & 22.40 & 0.604 & 3.990
            & 30.05 & \textbf{23.86} & 0.551 & \underline{4.005}
            & 26.82 & \underline{22.81} & 0.571 & 4.005 \\

        \hspace{.4em}SFT (Unpruned)
            & 24.70 & 22.86 & \underline{0.639} & \underline{4.005}
            & \underline{26.49} & \underline{22.50} & \textbf{0.620} & \underline{4.000}
            & \textbf{30.24} & 23.45 & \underline{0.563} & 4.000
            & \underline{26.98} & 22.80 & \underline{0.579} & \textbf{4.020} \\

        \hspace{.4em}\textbf{SFT (P2E, Ours)}
            & \textbf{24.92} & \textbf{22.98} & \textbf{0.646} & \textbf{4.010}
            & \textbf{26.68} & \textbf{22.55} & \textbf{0.620} & \textbf{4.005}
            & \underline{30.22} & \underline{23.48} & \textbf{0.568} & \textbf{4.010}
            & \textbf{27.21} & \textbf{22.87} & \textbf{0.582} & \textbf{4.020} \\
        \midline

        \rowcolor[gray]{0.9}
        \multicolumn{17}{l}{\textit{Z-Image-Turbo}} \\

        \hspace{.4em}w/o PE
            & 27.05 & 23.16 & 0.673 & 4.005
            & 28.38 & 23.05 & 0.656 & 4.000
            & 30.66 & 23.54 & 0.630 & 4.005
            & 28.48 & 23.16 & 0.637 & 4.010 \\

        \hspace{.4em}Raw PE
            & 27.09 & 23.11 & 0.677 & 4.015
            & 28.37 & \underline{23.11} & 0.660 & 4.005
            & 30.63 & \underline{23.63} & 0.635 & 4.000
            & 28.45 & \underline{23.18} & 0.648 & \textbf{4.020} \\

        \hspace{.4em}SFT (Unpruned)
            & \underline{27.13} & \underline{23.29} & \underline{0.680} & \textbf{4.020}
            & \underline{28.42} & 23.09 & \underline{0.681} & \textbf{4.015}
            & \textbf{30.87} & 23.62 & \underline{0.646} & \textbf{4.010}
            & \underline{28.82} & 23.16 & \underline{0.651} & \underline{4.015} \\

        \hspace{.4em}\textbf{SFT (P2E, Ours)}
            & \textbf{27.28} & \textbf{23.32} & \textbf{0.681} & \textbf{4.020}
            & \textbf{28.55} & \textbf{23.14} & \textbf{0.682} & \textbf{4.015}
            & \underline{30.86} & \textbf{23.68} & \textbf{0.648} & \textbf{4.010}
            & \textbf{28.89} & \textbf{23.22} & \textbf{0.653} & 4.010 \\
        \midline

        \rowcolor[gray]{0.9}
        \multicolumn{17}{l}{\textit{MOSS-Image}} \\

        \hspace{.4em}w/o PE
            & \underline{26.85} & 23.24 & 0.622 & 4.065
            & 25.63 & 22.07 & 0.604 & \textbf{4.015}
            & 30.09 & 23.18 & 0.523 & \textbf{4.005}
            & 26.76 & 22.36 & 0.560 & 4.010 \\

        \hspace{.4em}Raw PE
            & 26.28 & 23.28 & 0.635 & 4.065
            & 25.87 & 22.58 & 0.595 & 4.000
            & 30.06 & \textbf{23.50} & 0.527 & \textbf{4.005}
            & 26.72 & 22.80 & \underline{0.568} & 4.020 \\

        \hspace{.4em}SFT (Unpruned)
            & 26.69 & \underline{23.48} & \textbf{0.639} & \underline{4.070}
            & \textbf{26.73} & \underline{22.80} & \textbf{0.612} & \textbf{4.015}
            & \underline{30.12} & 23.43 & \underline{0.530} & 4.000
            & \underline{27.42} & \underline{23.02} & 0.565 & \textbf{4.025} \\

        \hspace{.4em}\textbf{SFT (P2E, Ours)}
            & \textbf{26.90} & \textbf{23.51} & \textbf{0.639} & \textbf{4.075}
            & \underline{26.53} & \textbf{22.83} & \underline{0.610} & 4.005
            & \textbf{30.22} & \underline{23.48} & \textbf{0.532} & \textbf{4.005}
            & \textbf{28.08} & \textbf{23.19} & \textbf{0.569} & \textbf{4.025} \\
        \midline

        \rowcolor[gray]{0.9}
        \multicolumn{17}{l}{\textit{BAGEL}} \\

        \hspace{.4em}w/o PE
            & 27.27 & 22.81 & 0.624 & 4.055
            & 27.01 & 22.46 & 0.603 & 3.980
            & 30.05 & \underline{23.31} & 0.571 & 4.000
            & 28.26 & \textbf{22.67} & 0.584 & \underline{4.035} \\

        \hspace{.4em}Raw PE
            & 27.46 & 22.90 & 0.625 & 4.055
            & \underline{27.34} & \textbf{22.51} & 0.618 & 3.990
            & 30.57 & 23.24 & 0.586 & \underline{4.005}
            & 27.95 & 22.63 & 0.598 & 4.005 \\

        \hspace{.4em}SFT (Unpruned)
            & \underline{27.82} & \textbf{22.97} & \underline{0.642} & \underline{4.100}
            & 27.12 & 22.36 & \textbf{0.642} & \textbf{4.010}
            & \underline{30.64} & 23.26 & \textbf{0.613} & \underline{4.005}
            & \textbf{28.56} & \underline{22.66} & \textbf{0.610} & \textbf{4.040} \\

        \hspace{.4em}\textbf{SFT (P2E, Ours)}
            & \textbf{28.03} & \underline{22.93} & \textbf{0.643} & \textbf{4.105}
            & \textbf{27.42} & \underline{22.49} & \underline{0.633} & \underline{3.995}
            & \textbf{30.79} & \textbf{23.45} & \underline{0.612} & \textbf{4.015}
            & \underline{28.47} & \underline{22.66} & \underline{0.603} & \underline{4.035} \\
    \end{tabular}%
    \vspace{-.8em}
\end{table*}

\section{Experiments}
\label{sec:exp}
In this section, we first describe the experimental setup and evaluate P2E across multiple visual generators. We then conduct systematic ablation studies to examine the effects of key hyperparameters and analyze model behaviors.

\subsection{Experimental Setups}
\paragraph{Training.}
We sample 60K high-quality images from LSDIR~\cite{li2023lsdir}, spanning four diverse categories: Landscape, Design, Portrait, and General. Following \autoref{sec:exp_setup_demo}, we construct paired dense captions and short, user-like prompts for SFT. We adopt Qwen3.5-9B~\cite{yang2025qwen3} as the PE backbone.
We fine-tune the PE model for $2$ epochs using AdamW~\cite{loshchilov2017decoupled} with a fixed learning rate of $1\times10^{-5}$. Training is performed on $8$ NVIDIA H20 GPUs with a per-device batch size of $16$ and a gradient accumulation step of $1$. For P2E, we generate $K=3$ nested pruning candidates targeting approximately $20\%$, $40\%$, and $60\%$ token removal, respectively\footnote{Due to the discrete nature of sentence-level pruning, actual deletion ratios may deviate from their targets. For instance, a candidate targeting $60\%$ removal may still remove no more than $40\%$ of the tokens.}, with a per-example deletion bound of $\tau=0.4$.

\paragraph{Evaluation.}
\label{sec:eval}
We curate a held-out test set of $800$ prompts, comprising $200$ prompts from each category. We assess PE performance across four different pretrained visual generators, including Qwen-Image-2.1-Turbo~\cite{zhao2026qwen}, Z-Image-Turbo~\cite{cai2025z}, MOSS-Image~\cite{ai2026moss}, and BAGEL~\cite{deng2025emerging}, using the four metrics established in \autoref{sec:exp_setup_demo}.

\subsection{Main Results}
To systematically evaluate the effectiveness of our proposed P2E method, we consider three representative baselines: direct generation without prompt enhancement (w/o PE), the pretrained PE model without fine-tuning (Raw PE), and conventional SFT using unpruned dense captions.

\paragraph{Effectiveness of P2E.}
As shown in \autoref{tab:main_results}, P2E improves both human preference and perceptual quality, demonstrating the benefits of refining supervision for prompt enhancement. By removing image-specific details weakly related to user intent while preserving explicit requirements and compositional structure, P2E provides more focused learning signals for meaningful visual enrichment.

\begin{figure*}[t]
    \centering
    \includegraphics[width=0.98\linewidth]{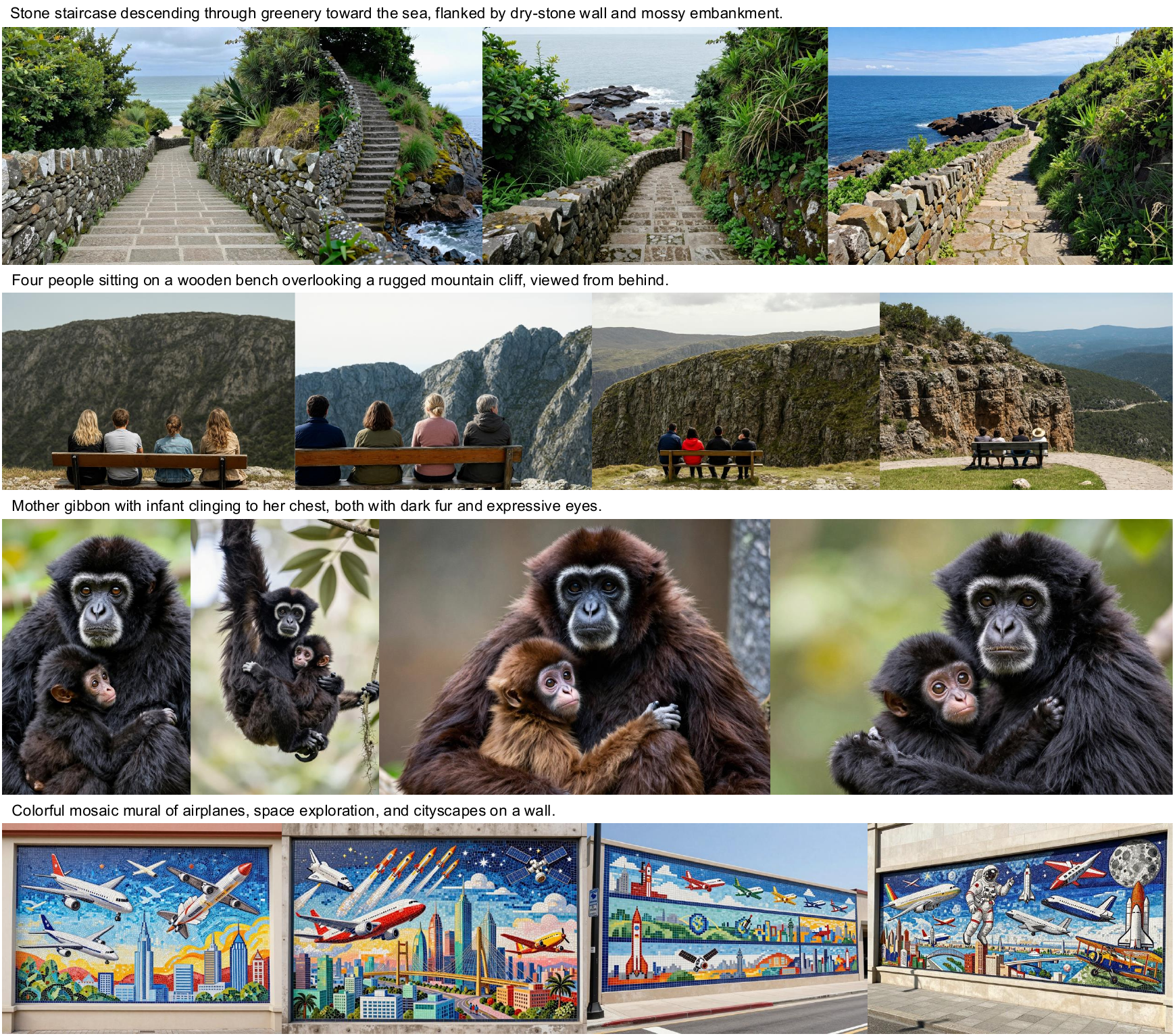}
    \vspace{-.3em}
    \caption{\textbf{Qualitative comparison of prompt enhancement methods.} From left to right: w/o PE, Raw PE, SFT (Unpruned), and SFT (P2E). The original prompt is shown above each row. P2E enhances structural coherence, spatial balance, and visual richness while faithfully preserving user intent, as reflected in clearer scene depth, more accurate subject attributes, and richer fine-grained details.}
    \vspace{-.8em}
    \label{fig:visual_compare}
\end{figure*}

The qualitative comparisons in \autoref{fig:visual_compare} further support these findings, showing more coherent scene structures, balanced spatial layouts, and richer visual details. In particular, P2E enhances spatial depth and subject placement in the landscape scenes, better preserves the specified dark fur of the gibbons, and integrates diverse mural elements into a cohesive composition. These results demonstrate that refined supervision improves compositional aesthetics and visual richness while maintaining fidelity to user intent.

\paragraph{Transferability Across Visual Generators.}
Beyond improving downstream generation quality, P2E also generalizes across different visual generators. As shown in \autoref{tab:main_results}, the same PE model trained with P2E-refined supervision generally outperforms conventional SFT across multiple generators, with particularly consistent gains in human preference. For instance, P2E increases HPS v2 from $26.98$ to $27.21$ on General with Qwen-Image-2.1-Turbo. With BAGEL, it improves HPS v2 from $27.12$ to $27.42$ and PickScore from $22.36$ to $22.49$ on Design. These results suggest that refining supervision helps the PE model learn transferable prompt enrichment capabilities, enabling its benefits to extend beyond a single downstream generator.

\paragraph{Risk of Fine-tuning Degradation.}
The results in \autoref{tab:main_results} also reveal that fine-tuning on unpruned dense captions carries a risk of underperforming zero-shot Raw PE. This may reflect overfitting when the available training data are \textit{insufficient} to capture the complexity of intent-preserving prompt enhancement. For instance, conventional SFT reduces HPS v2 from $27.34$ to $27.12$ on BAGEL Design and PickScore from $23.86$ to $23.45$ on Qwen-Image-2.1-Turbo Portrait. P2E mitigates this risk through supervision refinement. On BAGEL Design, it \textit{reverses} the degradation, achieving an HPS v2 score of $27.42$ and surpassing both Raw PE and conventional SFT. These findings suggest that effective fine-tuning requires supervision aligned with the goal of prompt enhancement. Reproducing image-specific details weakly related to user intent may offer limited value for this objective. P2E reduces such content while retaining explicit requirements and meaningful compositional information, providing more focused supervision for intent-aligned prompt enrichment.

\subsection{Ablation Studies}
We examine key P2E design choices through ablations using Qwen-Image-2.1-Turbo~\cite{zhao2026qwen}. All metrics in \autoref{tab:abla_K} and \autoref{tab:abla_tau} are averaged across the four evaluation categories described in \autoref{sec:eval}.

\paragraph{Candidate Granularity.}
We investigate the effect of candidate granularity by varying the number of pruning candidates $K$ while fixing the deletion bound at $\tau=0.4$. As shown in \autoref{tab:abla_K}, increasing $K$ from $3$ to $9$ yields no consistent improvement in either human preference or visual quality. This suggests that a finer discretization of removal ratios does not necessarily lead to more effective supervision. We therefore adopt $K=3$ as the default setting in all experiments, balancing downstream performance with the cost of offline candidate generation.

\paragraph{Deletion Bound.}
We examine the effect of the per-example deletion bound $\tau$ on downstream PE performance by varying $\tau\in\{0.2,0.4,0.6\}$ while fixing the number of pruning candidates at $K=9$. As shown in \autoref{tab:abla_tau}, $\tau=0.4$ achieves the best overall performance, suggesting a favorable balance between redundancy removal and the preservation of useful descriptive content in the supervision targets. These findings highlight the need to carefully control pruning intensity, as overly conservative pruning may retain image-specific details weakly related to user intent, whereas excessive pruning may weaken the supervision signal by discarding information needed to learn effective, intent-aligned prompt enrichment.

\paragraph{Semantic Preservation.}
To examine whether prompt enhancement preserves the semantics of the original short prompts, we evaluate the generated images on GenEval~\cite{ghosh2023geneval} against the original prompt requirements. As shown in \autoref{tab:geneval}, the P2E-trained PE model achieves a higher overall score than direct generation without PE. This improvement suggests that P2E provides useful visual elaborations that facilitate the faithful realization of the original compositional requirements, supporting its ability to enrich prompts while maintaining semantic fidelity.

\paragraph{Effectiveness with Dense Inputs.}
We further evaluate P2E on DPG-Bench~\cite{hu2024ella} to examine whether prompt enhancement remains beneficial when the input already provides rich descriptive information. Even for these detailed inputs, P2E improves the average DPG-Bench score over the w/o PE baseline (\autoref{tab:dpg_bench}), indicating improved adherence to their detailed requirements. These results suggest that P2E remains robust to relatively long, detailed prompts and can further improve generation performance, supporting its applicability across varying levels of prompt detail.

\begin{table}[t]
    \centering
    \tablestyle{4pt}{1.2}
    \footnotesize
    \caption{\textbf{Effect of candidate granularity.}
    We vary the number of pruning candidates $K$ by uniformly spacing target token removal ratios from $20\%$ to $60\%$, with intervals of $20\%$, $10\%$, and $5\%$ for $K=3$, $5$, and $9$, respectively. The deletion bound is fixed at $\tau=0.4$, with all other training settings unchanged.}
    \vspace{-.3em}
    \begin{tabular}{c|c|cc|cc}
         \multirow{2}{*}{\shortstack[c]{\textbf{\# of}\\\textbf{Candidates}}}
         & \multirow{2}{*}{\shortstack[c]{\textbf{Removal}\\\textbf{Interval (\%)}}}
         & \multicolumn{2}{c|}{\textbf{Preference}}
         & \multicolumn{2}{c}{\textbf{Visual Quality}} \\
         &
         & HPS v2$\uparrow$ & PicS$\uparrow$
         & MIQA$\uparrow$ & VQ-R1$\uparrow$ \\
        \shline
        $K=3$
            & 20
            & 27.26
            & \textbf{22.97}
            & 0.604
            & \textbf{4.011}
            \\
        $K=5$
            & 10
            & 27.14
            & 22.90
            & \textbf{0.606}
            & 4.003
            \\
        $K=9$
            & 5
            & \textbf{27.33}
            & 22.96
            & 0.603
            & 4.006
            \\
    \end{tabular}
    \vspace{-.8em}
    \label{tab:abla_K}
\end{table}

\begin{table}[t]
    \centering
    \tablestyle{4pt}{1.2}
    \footnotesize
    \caption{\textbf{Effect of deletion bound.}
    We evaluate different per-example deletion bounds $\tau$ under a fixed number of pruning candidates $K=9$, with target token removal ratios ranging from $20\%$ to $60\%$ at $5\%$ intervals.}
    \vspace{-.3em}
    \begin{tabular}{c|cc|cc}
         \multirow{2}{*}{\textbf{Deletion Bound}}
         & \multicolumn{2}{c|}{\textbf{Preference}}
         & \multicolumn{2}{c}{\textbf{Visual Quality}} \\
         & HPS v2$\uparrow$ & PicS$\uparrow$
         & MIQA$\uparrow$ & VQ-R1$\uparrow$ \\
        \shline
        Baseline ($\tau=0$)
            & 27.10
            & 22.90
            & 0.600
            & \textbf{4.006}
            \\
        $\tau=0.2$
            & 27.25
            & 22.84
            & 0.601
            & 3.995
            \\
        $\tau=0.4$
            & \textbf{27.33}
            & \textbf{22.96}
            & \textbf{0.603}
            & \textbf{4.006}
            \\
        $\tau=0.6$
            & 26.98
            & 22.56
            & 0.598
            & 3.993
            \\
    \end{tabular}
    \vspace{-.8em}
    \label{tab:abla_tau}
\end{table}

\section{Discussion and Limitation}
\paragraph{A Mirrored Perspective on P2E.}
Our work is fundamentally about data refinement, which lies at the heart of large model development. Constructing supervision from real images involves abstracting detailed visual descriptions into plausible user requests. This can be viewed as a many-to-one process, since distinct descriptions may share the same underlying intent. PE learns the reverse relation and therefore faces a one-to-many \textit{ambiguity}: a single request admits multiple plausible detailed elaborations. Yet conventional SFT uses the original dense caption as its supervision target, requiring the model to imitate both meaningful elaborations and incidental image-specific details. The resulting challenge is to identify which parts of this description provide useful supervision for enriching the synthesized request.

P2E approaches this challenge through a \textit{mirror process}. While PE expands a short prompt toward a detailed description, P2E starts from the paired dense caption and progressively removes sentences weakly related to the fixed prompt. This refinement reduces the target's reliance on incidental details of a particular visual realization. We conjecture that it thereby mitigates supervision ambiguity in PE's one-to-many mapping, strengthening the correspondence between user intent and supervision content and providing more effective and focused learning signals for meaningful, intent-aligned enrichment.

\paragraph{Limitation and Future Work.}
P2E relies on LLM judgments of sentence relevance. Future work could examine how LLMs of varying capabilities affect pruning effectiveness and whether the benefits of removing redundant supervision persist on substantially larger training datasets.

\section{Conclusion}
We revisit supervision design for prompt enhancement and uncover a counterintuitive finding: dense captions yielding larger generation gains do not necessarily provide more effective supervision. Motivated by this discrepancy, we introduce \textbf{Prune to Expand} (P2E), a text-only method that selectively removes redundant sentences under a per-example deletion bound while retaining explicit user requirements and informative visual structure. Under identical training configurations, P2E improves human preference and visual quality over conventional dense-caption SFT across diverse image categories and visual generators. More broadly, we advocate \textit{prune to expand} as a supervision design principle: giving models less redundant content to imitate can enable more effective learning and richer, intent-aligned prompt enrichment.

\begin{table}[t]
\caption{\textbf{Prompt-following evaluation on GenEval~\cite{ghosh2023geneval}.} We assess whether P2E-based prompt enhancement preserves the compositional requirements specified in the original short prompts.}
\label{tab:geneval}
\tablestyle{1.8pt}{1.2}
\begin{tabular}{l|c|cccccc}
\textbf{Method}
& \textbf{Overall}
& \begin{tabular}[c]{@{}c@{}}\textbf{Single}\\ \textbf{Obj.}\end{tabular}
& \begin{tabular}[c]{@{}c@{}}\textbf{Two}\\ \textbf{Obj.}\end{tabular}
& \textbf{Count.}
& \textbf{Colors}
& \textbf{Pos.}
& \begin{tabular}[c]{@{}c@{}}\textbf{Color}\\ \textbf{Attr.}\end{tabular} \\
\shline
w/o PE    &43.30  &83.75 &43.43 &42.19 &61.70 &9.50 &19.25 \\
SFT (P2E, Ours) &\textbf{74.23} &\textbf{98.12} &\textbf{85.10} &\textbf{71.56} &\textbf{81.12} &\textbf{55.25} &\textbf{54.25} \\
\end{tabular}
\end{table}

\begin{table}[t]
    \centering
    \tablestyle{2.3pt}{1.2}
    \footnotesize
    \caption{\textbf{Dense prompt evaluation on DPG-Bench~\cite{hu2024ella}.}
    We assess whether P2E remains effective when the input prompts already contain detailed descriptions.}
    \label{tab:dpg_bench}
    \vspace{-.3em}
    \begin{tabular}{l|c|ccccc}
        \textbf{Method}
        & \textbf{Avg}
        & \textbf{Global}
        & \textbf{Entity}
        & \textbf{Attribute}
        & \textbf{Relation}
        & \textbf{Other} \\
        \shline
        w/o PE
            &82.77 &80.55 &89.39 &88.47 &93.41 &85.60 \\
        SFT (P2E, Ours)
            &\textbf{84.97} &\textbf{83.28} &\textbf{91.53} &\textbf{90.80} &\textbf{94.41} &\textbf{88.00} \\
    \end{tabular}%
\end{table}

\section*{LLM Usage Statement}
In this work, we used GPT-6 Astra to assist with research code implementation and to polish the manuscript for clarity, readability, and linguistic consistency. Additionally, LLMs were used to construct synthetic data for training and evaluation. LLMs were not involved in conceiving the research ideas, developing the conceptual framework, formulating hypotheses, designing experiments, analyzing or interpreting experimental results, or drawing scientific conclusions. All AI-assisted code and textual revisions were carefully reviewed and verified by the authors to ensure their correctness and consistency with the intended scientific content. The authors take full responsibility for all aspects of this work, including its research contributions, synthetic data, experimental results, and final manuscript.

{
    \small
    \bibliographystyle{ieeenat_fullname}
    \bibliography{main}
}


\end{document}